\documentclass{article}
\usepackage{spconf,amsmath,epsfig}
\usepackage{amsmath, amssymb}
\usepackage[numbers]{natbib}
\usepackage{lineno}
\usepackage[linesnumbered,boxed,ruled,commentsnumbered]{algorithm2e}
\usepackage{xcolor}
\usepackage{hhline}
\usepackage{booktabs}
\usepackage{tabularx}
\usepackage{pifont}
\usepackage{mathrsfs}
\newcommand{\xmark}{\ding{55}}
\newcommand{\cmark}{\ding{51}}
\newcolumntype{Y}{>{\centering\arraybackslash}X}
\newcommand{\mediumsize}{\fontsize{8}{10}\selectfont}

\title{MANIFOLD-BASED SHAPLEY FOR SAR RECOGNIZATION NETWORK EXPLANATION}
\name{Xuran Hu\textsuperscript{1}, Mingzhe Zhu\textsuperscript{1, 2}, Yuanjing Liu\textsuperscript{1}, Zhenpeng Feng\textsuperscript{1}, Ljubi\v{s}a Stankovi\'c\textsuperscript{3}}
\address{\textsuperscript{1} School of Electronic Engineering, Xidian University, China \\ \textsuperscript{2} Kunshan Innovation Institute of Xidian University, China \\ \textsuperscript{3} Faculty of Electrical Engineering, University of Montenegro, Montenegro}

\begin{document}
%
\maketitle
\begin{abstract}
Explainable artificial intelligence (XAI) holds immense significance in enhancing the deep neural network's transparency and credibility, particularly in some risky and high-cost scenarios, like synthetic aperture radar (SAR). Shapley is a game-based explanation technique with robust mathematical foundations. However, Shapley assumes that model's features are independent, rendering Shapley explanation invalid for high dimensional models. This study introduces a manifold-based Shapley method by projecting high-dimensional features into low-dimensional manifold features and subsequently obtaining Fusion-Shap, which aims at (1) addressing the issue of erroneous explanations encountered by traditional Shap; (2) resolving the challenge of interpretability that traditional Shap faces in SAR recognization tasks.
\end{abstract}
\section{Introduction}
\label{sec:intro}

Synthetic aperture radar (SAR) is widely utilized in earth observation, electronic reconnaissance, and other fields due to its day-and-night and all-weather imaging capability. With the extraordinary ability of feature representation, deep neural networks (DNNs) are widely used in various SAR tasks, like object detection/localization, target identification, etc. However, due to the black-box nature of DNNs, it is difficult for human to understand DNN's decision-making logic. Simultaneously, \cite{feng2023analytical} demonstrates there are significant differences in DNN's decision-making processes between optical images and SAR images. These probably cause uncertainty in decision-making and conceal vulnerability of DNNs, particularly in critical domains such as military target reconnaissance. Therefore, network explanation is significant for evaluating system reliability and robustness.


 Shapley method \cite{shapley1953value, lundberg2017unified} is a commonly used network explanation that transforms the network explanation problem into an optimal allocation problem of network confidence. Specifically, the saliency map is obtained by calculating the marginal contribution of each input pixel to the network's confidence. Shapley is a typical local, ex-post, model-agnostic explanation that can explain any model. However, shapley method is usually based on an assumption that model's features are independent. This assumption is generally valid in low-dimensional models, but it often yields incorrect explanation when the network has high-dimensional features. When calculating the marginal contribution of a feature coalition, shapley method often provides feature coalition that do not conform to the manifold, rendering the calculated shapley value without practical significance.

Currently, researchers have begun to address this problem \cite{chen2023algorithms}. \cite{chang2018explaining} obtain more credible explanations by analyzing data distribution, \cite{anders2020fairwashing} solve the feature correlation problem through gradient methods, and \cite{aas2021explaining} attempt to extend the kernel method in shapley to obtain more accurate explanation estimates. \cite{frye2020shapley} proposed a shapley method that respects manifold distribution and provided two ways (unsupervised and supervised) to obtain interpretation estimations. Some efforts also aim to solve the manifold problem of explanation \cite{kwon2022weightedshap, albini2022counterfactual}. However, these methods are challenging to extend to high-dimensional data. With the development of generative networks, it has become possible to obtain reliable data manifold. This study utilizes generative adversarial networks (GAN) \cite{goodfellow2014generative} as a manifold deduction method and obtains fusion shapley by combining traditional and manifold shapley. The primary contributions of this paper are as follows: (1) We proposed a novel explanation method called Fusion-Shap, combining manifold and traditional Shap to obtain a reliable network explanation in SAR tasks. (2) We proposed a manifold-based shapley method that relies on obtaining a reliable manifold distribution through advanced generation networks.


\begin{figure}[!t]
	\centering
	\includegraphics[width=0.85\linewidth]{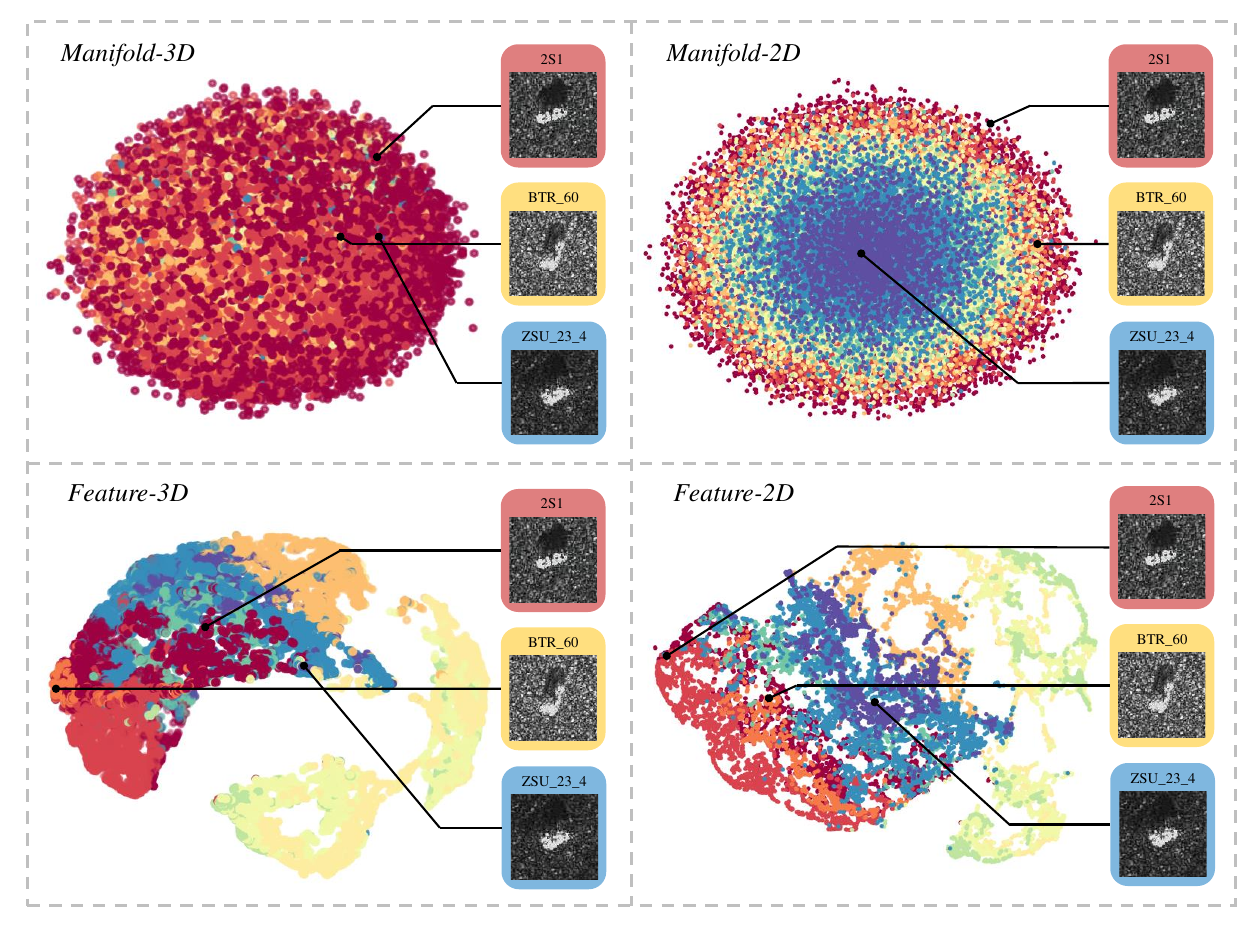}
	\caption{UMAP \cite{mcinnes2018umap-software} visualization of low-dimensional manifolds and high-dimensional features.}
	\label{fig_1}
\end{figure}

\section{Methodology}
\label{section:B}

As discussed in Section \ref{sec:intro}, traditional shapley method makes an assumption that features are independent, which, however, frequently does not hold. This assumption can result in misleading interpretations of high-dimensional features and consequently yield inaccurate network explanation. We employ StyleGAN to transform high-dimensional features into manifold to address this issue. This section details the implementation of Fusion-Shap. 


\subsection{Calculation of data manifold}\label{1.1}

We denote the network as $f$, with the sample to be explained represented as $I \in \mathbb{R}^{C \times W \times H}$, and the data manifold as $U \in \mathbb{R}^{1 \times L}$. We leverage the StyleGAN2 \cite{Karras2019stylegan2} framework to train the generator, denoted as $G$, which learns the underlying manifold structure and decodes the mapping, resulting in $I = G(U)$. Subsequently, we proceed to train the reconstructor, denoted as $R$, using Image2StyleGAN \cite{abdal2019image2stylegan} to obtain the encoding mapping: $U = R(I)$. 

%
%
%
%
Now, we have successfully established a mutual mapping relationship between high-dimensional data and low-dimensional manifolds, denoted as $I \Leftrightarrow U$.

\begin{figure*}[!t]
	\centering
	\includegraphics[width=0.9\textwidth]{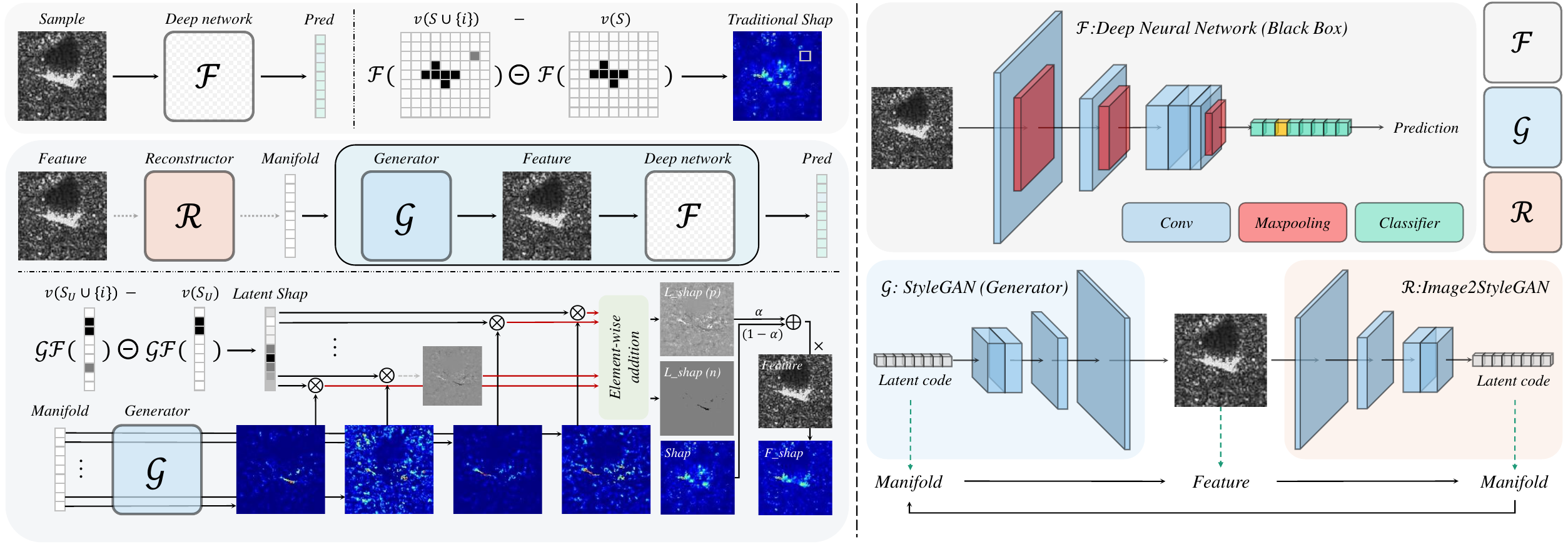}
	\caption{Left: Implementation of Fusion-shap; Right: The black-box model under interpretation:$F$. StyleGAN generators($G$) and Image2StyleGAN($R$), which enable the transformation of high-dimensional feature and low-dimensional manifolds.}
	\label{fig_2}
\end{figure*}

\subsection{Manifold-based Shapley}

In section \ref{1.1}, we have successfully mapped the high-dimensional feature to the low-dimensional manifold. This mapping allows us to compute the shapley value on the manifold directly. For a manifold comprising $L$ features, we can express formula (\ref{eq1}) as follows:

\begin{equation}\label{eq4}
	\begin{aligned}
		\phi_v(p) = & \sum_{S_U \subseteq L \backslash\{p\}} \frac{\left|S_U\right| !\left(l-\left|S_U\right|-1\right) !}{l !} \\
		& \cdot \left[v\left(S_U \cup\{p\}\right)-v\left(S_U\right)\right].
	\end{aligned}
\end{equation}

In this equation, $S_U$ represents a subset $S_U \subseteq L=\{1,2, \ldots, l\}$ of the feature set $L$, while $S_U \subseteq L \backslash\{p\}$ indicates that $S_U$ is a subset of $L$ and does not include feature $p$. The symbol $v$ denotes the value function. For a specific sample denoted as $u^\prime$, the function $v$ can be expressed as:

\begin{equation}
	v\left(S_U\right)=E\left[f(g(U)) \mid U_S=U_S^{\prime}\right].
\end{equation}

\subsection{Shapley mapping}

We have successfully established a mapping between the high-dimensional feature, $I$, and the low-dimensional manifold, $U$. However, the shapley value represents feature importance and the challenge now lies in determining how to mapping this importance effectively. 


We adopte a gradient-based approach to assess this importance mapping. First, define $U_p$ as an element within the position-encoded $p$-index manifold. It is intuitive to understand that slight perturbations in $U_p$ will induce changes in the high-dimensional features represented by $I$. We define this mapping as $\gamma$, unit-importance-mapping, which can be expressed mathematically as follows:

\begin{equation}\label{eq8}
	\phi_v\left(I^p\right)=\gamma \left[\phi_v\left(U_p\right)\right].
\end{equation}

In this equation, $I$ represents an image, a coalition of pixels. Formula (\ref{eq8}) can be interpreted as mapping the shapley value from the low-dimensional manifold to the high-dimensional feature space for each feature:

\begin{equation}\label{eq9}
	\phi_v\left(I_{c, w, h}^p\right)=K \cdot \varphi_{c, w, h}^p \cdot \phi_v\left(U_p\right)=K \cdot \frac{\partial I_{c, w, h}}{\partial U_p} \cdot \phi_v\left(U_p\right).
\end{equation}

To preserve the shapley characteristics within the shapley methods, we express the shapley value $\phi_v\left(U_p\right)$ of $U_p$ as a weighted sum of the individual shapley values $\phi_v\left(I_{c, w, h}^p\right)$ within the high-dimensional feature space $I$:

\begin{equation}\label{eq10}
	\phi_v\left(U_p\right)=\sum_c \sum_w \sum_h \omega_{c, w, h}^p \cdot \phi_v\left(I_{c, w, h}^p\right).
\end{equation}

Combining Formulas (\ref{eq9}) and (\ref{eq10}):

\begin{equation}
	\phi_v\left(I_{c, w, h}^p\right)=\frac{\partial I_{c, w, h}}{\partial U_p} \cdot \frac{\phi_v\left(U_p\right)}{C \times W \times H}.
\end{equation}

For all elements in manifold $U$:

\begin{equation}
	\phi_v^{M}\left(I_{c, w, h}\right)=\sum_p \phi_v\left(I_{c, w, h}^p\right)=K \sum_p \frac{\partial I_{c, w, h}}{\partial U_p} \cdot \phi_v\left(U_p\right)
\end{equation}

Now, we have obtained the saliency map denoted as $M_{\text {manifold }}=\phi_v^{M}\left(I_{c, w, h}\right)$.


\subsection{Shapley Fusion}

Manifold-based successfully addresses the issue of traditional Shapely method's overlooking on feature interdependence. Nevertheless, the manifold dimension somewhat constrains this method, and altering the manifold dimension necessitates relearning the mapping relationship between high-dimensional features and low-dimensional manifold—a process that incurs substantial computational costs. To tackle this problem, we propose a hybrid approach by reintegrating the feature-independent traditional shapley method into the manifold shapley method:

\begin{equation}\label{eq13}
	M_{fusion }=\alpha M_{manifold }+(1-\alpha) M_{traditional }.
\end{equation}

The fusion coefficient $\alpha \in [0, 1]$ is determined with the objective of minimizing average drop in confidence, which can be expressed mathematically as:

\begin{equation}
	\operatorname{argmin} \sum_{t=1}^C\left\{\operatorname{hardmax}\left[f(I)_t\right]-f\left(M_{fusion  }\otimes I \right)_t\right\},
\end{equation} among them, $\otimes$ stands for Hadamard product. The set $t=\{1,2 \ldots, C\}$ signifies that the network encompasses a total of $C$ confidence categories, with $t$ representing the Top-1 confidence category assigned to the input $I$. Utilizing equation (\ref{eq13}), we can derive the fusion coefficient $\alpha$ and subsequently compute Fusion-Shap, denoted as $M_{fusion}$, elements in saliency map are Fusion-shapley value $\phi_v^{F}$.

\section{Experimental Results}
\label{section:C}

\subsection{Experiment Settings}

\noindent \textbf{Dataset:} This paper employs MSTAR dataset.

\noindent \textbf{Evaluation Metrics:} We conducted qualitative and quantitative assessments of explanation methods. First, drawing upon prior research \cite{deng2023understanding}, we ascertain whether these methods adhere to a robust mathematical foundation, explicitly addressing the three primary criteria of interpretability. Secondly, we introduce quantitative metrics, fidelity and sensitivity, to mathematically evaluate the performance of explanation methods \cite{yeh2019fidelity}. 

%
%

\begin{figure}[!t]
	\centering
	\includegraphics[width=0.9\linewidth]{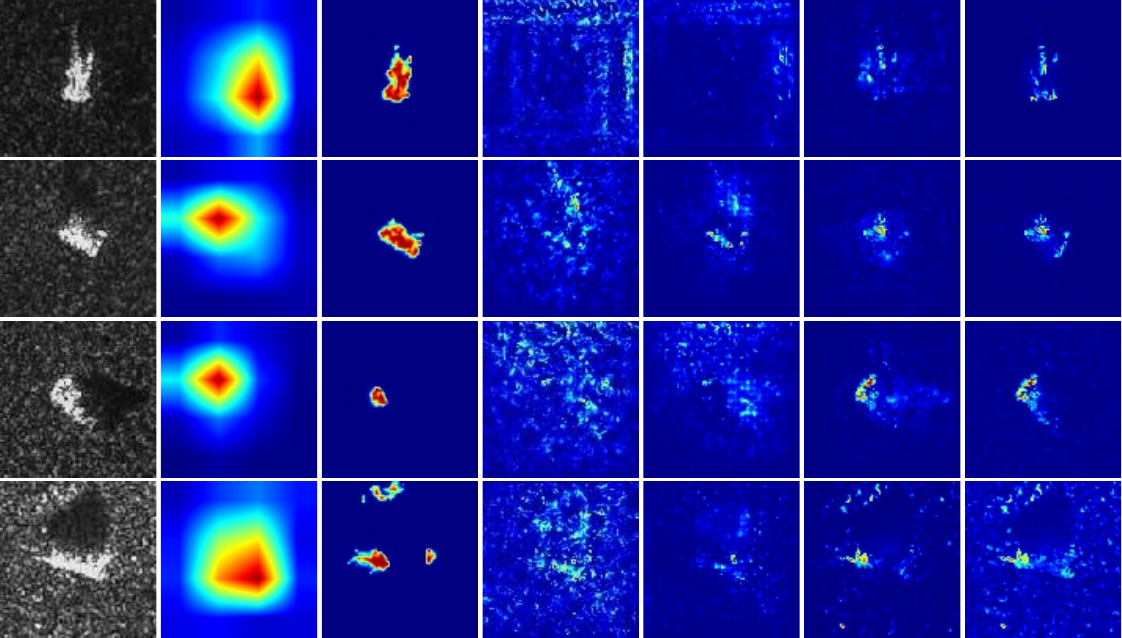}
	\caption{Results visualization. Columns from left to right: Origin image, Grad CAM, LRP, IG, SG, SHAP and F-SHAP.}
	\label{fig_3}
\end{figure}


\subsection{Visualization and Subjective Evaluation}

In contrast to optical images and human intuitive perception, it is more difficult to understand DNNs' decision-making logic on SAR images due to the intricate nature of SAR imaging mechanism. SAR recognition network's decision-making is not solely contingent upon the target area; interference spots and shadow regions also holds significance in network decision-making. Figure \ref{fig_3} illustrates visualization results of Grad CAM \cite{selvaraju2017grad}, LRP \cite{bach2015pixel}, IG \cite{sundararajan2017axiomatic}, SG \cite{smilkov2017smoothgrad}, SHAP \cite{lundberg2017unified} and Fusion-SHAP. Table \ref{tab1} shows the results of subjective evaluation (explanation validity from lowest to highest: 1-10).

\subsection{Axiomatic Validation}

Taylor Interactions \cite{deng2023understanding} considers various explanation methods by aggregating the individual effects $\varphi(k)$ and the interaction $I(k)$ of characteristic effects, each governed by distinct rules. Subsequently, it introduces three criteria for assessing the attribution method's reliability: Low Approximation Error (LAE), No Unrelated Allocation (NUA) and Complete Allocation (CA). The results of the attribution validation are presented in Table \ref{tab1}. 

%
%


\begin{table}[!t]
	\caption{Axioms, Sen., Infd. and subjective evaluation.}
	\label{tab1}
	\centering
	\mediumsize
	\renewcommand{\arraystretch}{1.3}
	\begin{tabular}{lcccccc}
		\toprule
		& \multicolumn{3}{c}{\textbf{Axiomatic}} & \multicolumn{2}{c}{\textbf{Metric}} & \textbf{Sub.} \\
		\cmidrule(lr){2-4} \cmidrule(lr){5-6}
		& \textbf{LAE} & \textbf{NUA} & \textbf{CA} & \textbf{INFD} & \textbf{SEN} & \\
		\midrule
		Grad\cite{shrikumar2017learning} & \xmark & \cmark & \cmark & 0.0423 & 269.31 & 2.15 \\
		GradCAM\cite{selvaraju2017grad} & \xmark & \cmark & \cmark & 3.5e-6 & 2.4124 & 1.00 \\
		LRP\cite{bach2015pixel} & \cmark & \xmark & \cmark & 0.0552 & 48.120 & 5.10 \\
		I-Grad\cite{sundararajan2017axiomatic} & \cmark & \cmark & \cmark & 0.0697 & 3.4425 & 2.90 \\
		S-Grad\cite{smilkov2017smoothgrad} & \cmark & \cmark & \cmark & 0.0079 & 4.0212 & 4.15 \\
		Shapley\cite{lundberg2017unified} & \cmark & \cmark & \cmark & 0.0009 & 3.4156 & 5.70 \\
		F-SHAP & \cmark & \cmark & \cmark & 7.1e-5 & 2.1285 & 5.95 \\
		\bottomrule
	\end{tabular}
\end{table}

(\ref{eq4}) compute the shapley values for high-dimensional features and low-dimensional manifolds. Both of these components notably satisfy the three computed aforementioned characteristics. The contribution of each feature within manifold can be decomposed as follows:

\begin{equation}
	\phi_v^p=\sum_{k \in \Omega_p} \varphi(k)+\sum_{\left|S_U\right|>1, p \in S_U} \sum_{k \in \Omega_S} \frac{1}{\left|S_U\right|} I(k).
\end{equation}

Fusion-Shap can be viewed as the redistribution of shapley values through manifold information. Since $\sum \phi_v(i)=\sum \phi_v(p)$, this distribution still adheres to the shapley value properties. To illustrate this more intuitively, the shapley value of feature $U_p$ within manifold $U$ is reassigned to the high-dimensional space $I$, corresponding to the cumulative shapley values of multiple high-dimensional features. We reasonably hypothesize that the shapley value of feature $U_p$ in manifold $U$ is mapped to the high-dimensional space and allocated to two or more features. Thus, we can express Fusion-Shap in the form of Taylor interaction:

\begin{equation}
	\begin{aligned}
		\phi_v^F(i)= & (1-\alpha) \phi_v(i)+\alpha \phi_v^M(i) \\
		= & (1-\alpha) \phi_v(i)+\alpha \sum_p K_p \phi_v(p) \\
		= & (1-\alpha)\left[\sum_{k \in \Omega_i} \varphi(k)+\sum_{|S|>1, i \in S} \sum_{k \in \Omega_S} \frac{1}{\left|S_U\right|} I(k)\right] \\
		& +\alpha \sum_p K_p \left[\sum_{k^{\prime} \in \Omega_p} \varphi\left(k^{\prime}\right)+\sum_{ p \in S_U} \sum_{k^{\prime} \in \Omega_S} \frac{1}{\left|S_U\right|} I\left(k^{\prime}\right)\right] \\
		= & (1-\alpha) \sum_{k \in \Omega_i} \varphi(k)+\sum_{i \in S^{\prime}} \sum_{k \in \Omega_S^{\prime}} \frac{1}{\left|S^{\prime}\right|} I(k) \\
		& +(1-\alpha) \sum_{|S|>1, i \in S \backslash S^{\prime}} \sum_{k \in \Omega_S} \frac{1}{\left|S \backslash S^{\prime}\right|} I(k).
	\end{aligned}
\end{equation}

This Taylor interaction formulation satisfies three axioms, substantiating the effectiveness of Fusion Shap.

\begin{figure}[!t]
	\centering
	\includegraphics[width=\linewidth]{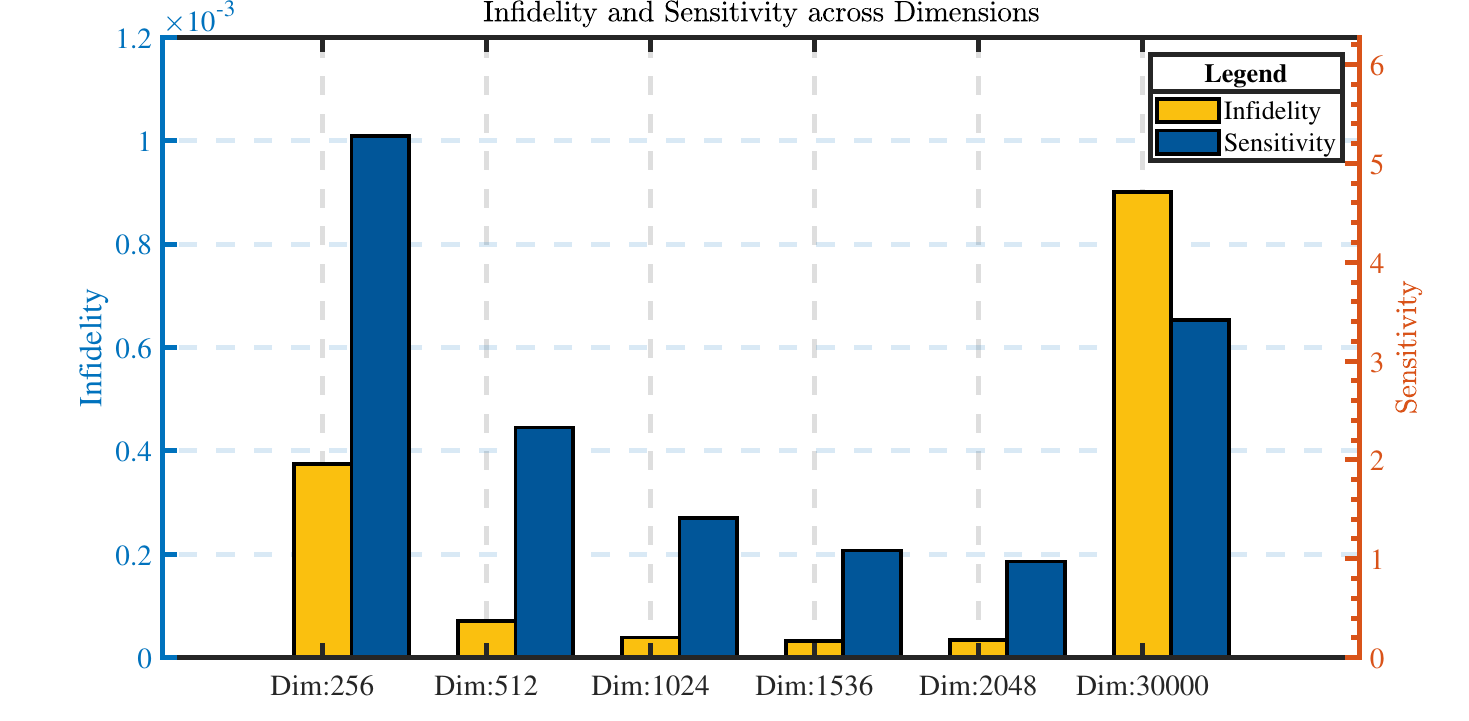}
	\caption{Infidelity and sensitivity vary across different dimensions of manifold. }
	\label{fig_4}
\end{figure}

\subsection{Explanation Sensitivity and Infidelity}



We introduce fidelity and sensitivity \cite{yeh2019fidelity} for evaluating model performance. For a given method $\psi(f, I)$, we consider a given meaningful perturbation $P$ with probability distribution $\mu_P$. Infidelity of $\psi$ can be defined as:

\begin{equation}
	\operatorname{INFD}(f, I, \psi)=E_{P \sim \mu_P}\left[P^T \psi(f, x)-(f(X)-f(x-P))^2\right],
\end{equation} and the sensitivity index can be expressed as:

\begin{equation}
	\operatorname{SEN}(f, I, \psi, r)=\max _{\|I^{\prime}-I\| \leq r}\|\psi(f, I^{\prime})-\psi(f, I)\|,
\end{equation} among them, parameter $r$ denotes the radius of the domain or field under consideration. 

Table \ref{tab1} presents the results of infidelity and sensitivity. F-shap meets the axiom verification criteria and achieves the best infidelity, sensitivity, and subjective evaluation metrics.

\subsection{Manifold Dimension and Explanation}
We have examined several manifold dimension configurations and calculated the fidelity and sensitivity of network explanations. Figure \ref{fig_4} showcase the manifold dimensions and model reliability. It is evident that increasing the manifold dimension reduces model's infidelity and sensitivity.



\section{Conclusions}
\label{section:F}

This study introduces a shapley-based method for elucidating SAR recognization networks. By combining low-dimensional manifold shap and high-dimensional feature shap, this approach rectifies the inherent assumption of feature independence. Simultaneously, we introduce shapley mapping to achieve the transformation between manifold shap and original shap. Experimental results confirm the efficacy of our method in terms of visualization, subjective evaluation, axiom validation, and infidelity and sensitivity assessment.

\bibliographystyle{IEEEbib}
\bibliography{strings,refs}

\begin{thebibliography}{10}

\bibitem{feng2023analytical}
Zhenpeng Feng, Hongbing Ji, Milo{\v{s}} Dakovi{\'c}, Mingzhe Zhu, and
  Ljubi{\v{s}}a Stankovi{\'c},
\newblock ``Analytical interpretation of the gap of cnn’s cognition between
  sar and optical target recognition,''
\newblock {\em Neural Networks}, vol. 165, pp. 982--986, 2023.

\bibitem{shapley1953value}
Lloyd~S Shapley et~al.,
\newblock ``A value for n-person games,''
\newblock 1953.

\bibitem{lundberg2017unified}
Scott~M Lundberg and Su-In Lee,
\newblock ``A unified approach to interpreting model predictions,''
\newblock {\em Advances in neural information processing systems}, vol. 30,
  2017.

\bibitem{chen2023algorithms}
Hugh Chen, Ian~C Covert, Scott~M Lundberg, and Su-In Lee,
\newblock ``Algorithms to estimate shapley value feature attributions,''
\newblock {\em Nature Machine Intelligence}, pp. 1--12, 2023.

\bibitem{chang2018explaining}
Chun-Hao Chang, Elliot Creager, Anna Goldenberg, and David Duvenaud,
\newblock ``Explaining image classifiers by counterfactual generation,''
\newblock in {\em International Conference on Learning Representations}, 2018.

\bibitem{anders2020fairwashing}
Christopher Anders, Plamen Pasliev, Ann-Kathrin Dombrowski, Klaus-Robert
  M{\"u}ller, and Pan Kessel,
\newblock ``Fairwashing explanations with off-manifold detergent,''
\newblock in {\em International Conference on Machine Learning}. PMLR, 2020,
  pp. 314--323.

\bibitem{aas2021explaining}
Kjersti Aas, Martin Jullum, and Anders L{\o}land,
\newblock ``Explaining individual predictions when features are dependent: More
  accurate approximations to shapley values,''
\newblock {\em Artificial Intelligence}, vol. 298, pp. 103502, 2021.

\bibitem{frye2020shapley}
Christopher Frye, Damien de~Mijolla, Tom Begley, Laurence Cowton, Megan
  Stanley, and Ilya Feige,
\newblock ``Shapley explainability on the data manifold,''
\newblock in {\em International Conference on Learning Representations}, 2020.

\bibitem{kwon2022weightedshap}
Yongchan Kwon and James~Y Zou,
\newblock ``Weightedshap: analyzing and improving shapley based feature
  attributions,''
\newblock {\em Advances in Neural Information Processing Systems}, vol. 35, pp.
  34363--34376, 2022.

\bibitem{albini2022counterfactual}
Emanuele Albini, Jason Long, Danial Dervovic, and Daniele Magazzeni,
\newblock ``Counterfactual shapley additive explanations,''
\newblock in {\em Proceedings of the 2022 ACM Conference on Fairness,
  Accountability, and Transparency}, 2022, pp. 1054--1070.

\bibitem{goodfellow2014generative}
Ian Goodfellow, Jean Pouget-Abadie, Mehdi Mirza, Bing Xu, David Warde-Farley,
  Sherjil Ozair, Aaron Courville, and Yoshua Bengio,
\newblock ``Generative adversarial nets,''
\newblock {\em Advances in neural information processing systems}, vol. 27,
  2014.

\bibitem{mcinnes2018umap-software}
Leland McInnes, John Healy, Nathaniel Saul, and Lukas Grossberger,
\newblock ``Umap: Uniform manifold approximation and projection,''
\newblock {\em The Journal of Open Source Software}, vol. 3, no. 29, pp. 861,
  2018.

\bibitem{Karras2019stylegan2}
Tero Karras, Samuli Laine, Miika Aittala, Janne Hellsten, Jaakko Lehtinen, and
  Timo Aila,
\newblock ``Analyzing and improving the image quality of {StyleGAN},''
\newblock in {\em Proc. CVPR}, 2020.

\bibitem{abdal2019image2stylegan}
Rameen Abdal, Yipeng Qin, and Peter Wonka,
\newblock ``Image2stylegan: How to embed images into the stylegan latent
  space?,''
\newblock in {\em Proceedings of the IEEE/CVF international conference on
  computer vision}, 2019, pp. 4432--4441.

\bibitem{deng2023understanding}
Huiqi Deng, Na~Zou, Mengnan Du, Weifu Chen, Guocan Feng, Ziwei Yang, Zheyang
  Li, and Quanshi Zhang,
\newblock ``Understanding and unifying fourteen attribution methods with taylor
  interactions,''
\newblock {\em arXiv preprint arXiv:2303.01506}, 2023.

\bibitem{yeh2019fidelity}
Chih-Kuan Yeh, Cheng-Yu Hsieh, Arun Suggala, David~I Inouye, and Pradeep~K
  Ravikumar,
\newblock ``On the (in) fidelity and sensitivity of explanations,''
\newblock {\em Advances in Neural Information Processing Systems}, vol. 32,
  2019.

\bibitem{selvaraju2017grad}
Ramprasaath~R Selvaraju, Michael Cogswell, Abhishek Das, Ramakrishna Vedantam,
  Devi Parikh, and Dhruv Batra,
\newblock ``Grad-cam: Visual explanations from deep networks via gradient-based
  localization,''
\newblock in {\em Proceedings of the IEEE international conference on computer
  vision}, 2017, pp. 618--626.

\bibitem{bach2015pixel}
Sebastian Bach, Alexander Binder, Gr{\'e}goire Montavon, Frederick Klauschen,
  Klaus-Robert M{\"u}ller, and Wojciech Samek,
\newblock ``On pixel-wise explanations for non-linear classifier decisions by
  layer-wise relevance propagation,''
\newblock {\em PloS one}, vol. 10, no. 7, pp. e0130140, 2015.

\bibitem{sundararajan2017axiomatic}
Mukund Sundararajan, Ankur Taly, and Qiqi Yan,
\newblock ``Axiomatic attribution for deep networks,''
\newblock in {\em International conference on machine learning}. PMLR, 2017,
  pp. 3319--3328.

\bibitem{smilkov2017smoothgrad}
Daniel Smilkov, Nikhil Thorat, Been Kim, Fernanda Vi{\'e}gas, and Martin
  Wattenberg,
\newblock ``Smoothgrad: removing noise by adding noise,''
\newblock {\em arXiv preprint arXiv:1706.03825}, 2017.

\bibitem{shrikumar2017learning}
Avanti Shrikumar, Peyton Greenside, and Anshul Kundaje,
\newblock ``Learning important features through propagating activation
  differences,''
\newblock in {\em International conference on machine learning}. PMLR, 2017,
  pp. 3145--3153.

\end{thebibliography}

\end{document}